\documentclass[11pt,a4paper]{article}
\usepackage[hyperref]{acl2021}
\usepackage{times}
\usepackage{latexsym}

\usepackage{microtype}
\usepackage{url}
\usepackage{color}
\usepackage{makecell}
\usepackage{array}
\usepackage{colortbl}
\usepackage{arydshln}
\usepackage{booktabs}
\usepackage{amsmath}
\usepackage{float}
\usepackage{graphicx}
\usepackage{tikz}
\usepackage{subfigure} 
\usepackage{multirow}
\usepackage{amsfonts}
\aclfinalcopy 

\definecolor{SpeakerAColor}{RGB}{60, 135, 196}
\definecolor{SpeakerBColor}{RGB}{208, 36, 101}

\title{Dialogue-oriented Pre-training}

\author{Yi Xu\textsuperscript{1,2,3}, Hai Zhao\textsuperscript{1,2,3,\thanks{\ \ Corresponding author. This paper has been accepted by ACL 2021 Findings. This paper was partially supported by National Key Research and Development Program of China (No. 2017YFB0304100), Key Projects of National Natural Science Foundation of China (U1836222 and 61733011), Huawei-SJTU long term AI project, Cutting-edge Machine Reading Comprehension and Language Model. This work was supported by Huawei Noah's Ark Lab. } } \\
\textsuperscript{1} Department of Computer Science and Engineering, Shanghai Jiao Tong University\\
\textsuperscript{2} Key Laboratory of Shanghai Education Commission for Intelligent Interaction\\
and Cognitive Engineering, Shanghai Jiao Tong University, Shanghai, China\\
\textsuperscript{3}MoE Key Lab of Artificial Intelligence, AI Institute, Shanghai Jiao Tong University\\
\texttt{xuyi\_2019@sjtu.edu.cn,zhaohai@cs.sjtu.edu.cn}\\
}
\date{}
\begin{document}
\maketitle
\begin{abstract}
Pre-trained language models (PrLM) has been shown powerful in enhancing a broad range of downstream tasks including various dialogue related ones. However, PrLMs are usually trained on general plain text with common language model (LM) training objectives, which cannot sufficiently capture dialogue exclusive features due to the limitation of such training setting, so that there is an immediate need to fill the gap between a specific dialogue task and the LM task. As it is unlikely to collect huge dialogue data for dialogue-oriented pre-training, in this paper, we propose three strategies to simulate the conversation features on general plain text. Our proposed method differs from existing post-training methods that it may yield a general-purpose PrLM and does not individualize to any detailed task while keeping the capability of learning dialogue related features including speaker awareness, continuity and consistency. The resulted Dialog-PrLM is fine-tuned on three
public multi-turn dialogue datasets and helps achieve significant and consistent improvement over the plain PrLMs. 
\end{abstract}

\section{Introduction}
Recently, pre-trained language models (PrLMs) have shown impressive improvements for various downstream NLP tasks \citep{zhou-etal-2020-limit,ouyang2020dialogue,zhang2021advances,radford2018improving,yang2019xlnet,zhang2020SemBERT,clark2020electra,9354025}, including the response selection task for multi-turn dialogues, which takes a dialogue history as input and aims to select a most suitable response from a collection of answers \citep{zhou2016multi,wu-etal-2017-sequential,zhou2018multi,zhu-etal-2018-lingke,zhang-etal-2018-modeling,tao2019multi,gu2019interactive}. 
\begin{table}
\renewcommand\tabcolsep{2pt}
\footnotesize
    \centering
    \begin{tabular}{l}
   \toprule
       \textbf{Dialogue History:} \\
A: \textit{Could you please get me a train?} \\
B: \textit{Sure I can help you find a train.} \\
B: \textit{where are you coming from?} \\
 B: \textit{What time do you need to leave by?} \\
 A: \textit{I am leaving Leicester and I need to leave by 20:30.} \\
B: \textit{What is your destination and day of travel?}\\
A: \textit{Cambridge and on friday.} \\
A: \textit{Can I just get the travel time for the train? Thanks!} \\
\hline
 \textbf{Response:} \\
B: \textit{The first train leaving after 20:30 is
142 21:09 and the} \\ 
\textit{travel time is 105 minutes.}\\
    \toprule
    \end{tabular}
    \caption{ \label{tab:dialogue_case}A multi-turn dialogue example with interleaved or continuous utterances
between two speakers.
    }
\end{table}

Pre-training tasks of all these PrLMs almost concentrate on two aspects: token prediction and sentence relation prediction. For example, the genetic BERT model \citep{devlin-etal-2019-bert} uses masked language modeling (MLM) and next sentence prediction (NSP) objectives; ALBERT \citep{Lan2020ALBERT:} predicts sentence order rather than NSP; ELECTRA \citep{clark2020electra} transfers MLM into a generating and then discriminating process like GAN \citep{goodfellow2014generative}.
However, these tasks are just devoted to incorporating token-level and sentence-level semantic information into embeddings, and cannot be sufficiently compatible with its dialogue-oriented characteristics.

Table \ref{tab:dialogue_case} shows a multi-turn dialogue example. Compared with plain text, the utterance turn and speaker role keep shift as a conversation goes on, and the next utterance should keep continuous and consistent with the context. Besides, the two speakers may not follow strict shift rules, and one speaker may continuously shoot multiple utterances. Although some existing works have noticed such nonlinear nature of multi-turn dialogues, they are limited to conducting post-training or pre-training in a specific domain and do not provide general-purpose dialogue-oriented PrLMs to fundamentally solve this problem \citep{xu2021learning,whang2021ums,wolf2019transfertransfo,zhang2019dialogpt,henderson-etal-2020-convert,bao-etal-2020-plato}. 


In this work, we make the first attempt to train a general-purpose dialogue-oriented PrLM. However, such a PrLM should be trained on huge dialogue data, which is hard to collect. Thus we propose three novel pre-training strategies (i.e., Insertion, Deletion, Replacement), so that we facilitate plain text originally for common PrLM training to simulate dialogue-like features. The resulted model, we denote as Dialog-PrLM, then is capable of effectively learning speaker awareness, continuity and consistency in a general way. Especially, for the convenient use of the downstream dialogue tasks, we introduce a special token $\texttt{[SOT]}$ before each utterance to tell that it is a start of a turn and learn from these three strategies. These targeted pre-training tasks enable $\texttt{[SOT]}$ to better represent each context utterance. We mimic dialogue-related features on conventional plain text, which can bring up the possibility that similar techniques could be adopted in other domains not only for dialogues. 



Our pre-trained Dialog-PrLM is fine-tuned on three multi-turn dialogue response selection benchmarks, and obtains significant and consistent improvements over the plain PrLMs.
\section{Related Work}
For multi-turn dialogue response selection task, earlier works conduct single-turn match, which concatenates all the utterances in the history dialogue or just considers the last one to match with the candidate response \citep{lowe-etal-2015-ubuntu,kadlec2015improved,10.1145/2911451.2911542,tan2015lstm,10.5555/3060832.3061030,wang-jiang-2016-learning}. Recently, existing works tend to model the interaction between each dialogue utterance and the response, which usually adopt the encoding-matching-aggregation paradigm \citep{zhou2016multi,wu-etal-2017-sequential,zhang-etal-2018-modeling,zhou2018emotional,zhou2018multi,tao2019multi,yuan2019multi}. 
After encoding, distinct matching networks generate features for each utterance which are usually passed to GRU \citep{cho-etal-2014-learning} for aggregating into a final matching score.
Besides, some works adopt topic information \citep{xing2017topic,wu2018response,xu2021topic} or conversation disentanglement to select the proper response \citep{jia-etal-2020-multi,wang-etal-2020-response}. 

There are more and more practice using powerful PrLMs as the model encoder \citep{zhang2021retro,zhang2019dcmn+,zhu2020dual} like BERT \citep{devlin-etal-2019-bert}, RoBERTa \citep{liu2019roberta}, ALBERT \citep{Lan2020ALBERT:} and ELECTRA \citep{clark2020electra}.  Considering task domain difference from the general corpus for PrLM pre-training, recent studies start to conduct post-training on target multi-turn dialogue datasets to incorporate in-domain knowledge \citep{whang2020domain,lu2020improving,gu2020speaker,xu2021learning,whang2021ums}. \citet{whang2020domain} conduct post-training of MLM and NSP tasks as BERT.  
Rather than using the same tasks as PrLMs, \citet{xu2021learning} and \citet{whang2021ums} both considers auxiliary tasks through post-training to enhance response selection.

Although the PrLMs which are trained on plain text have learned contextual semantic representation from token-level or sentence-level pre-training tasks like MLM, NSP, they all do not consider dialogue related features like speaker role, continuity and consistency. Despite some existing works  \citep{xu2021learning,whang2021ums} considers that when conducting post-training, they are limited to a specific domain. \citep{wolf2019transfertransfo,zhang2019dialogpt,henderson-etal-2020-convert,bao-etal-2020-plato} train on open-domain conversational data like Reddit for response selection or generation tasks, but they are limited to original pre-training tasks on plain text and ignore the dialogue related features. Besides, \citet{wu-etal-2020-tod} and \citet{li2020task} conduct task-specific training on collected dialogue corpora, but they also suffer from biased and limited amount of  dialogue data.

Different from all the previous studies, we still make an attempt in obtaining a general-purpose PrLM but not aiming at any specific tasks like post-training methods. Meanwhile, our proposed dialogue-oriented pre-training enables the resulted PrLMs to especially capture dialogue related features in a general way.   


\section{Dialogue-oriented Pre-training}
\begin{figure*}[ht]
\centering
\includegraphics[scale=0.58]{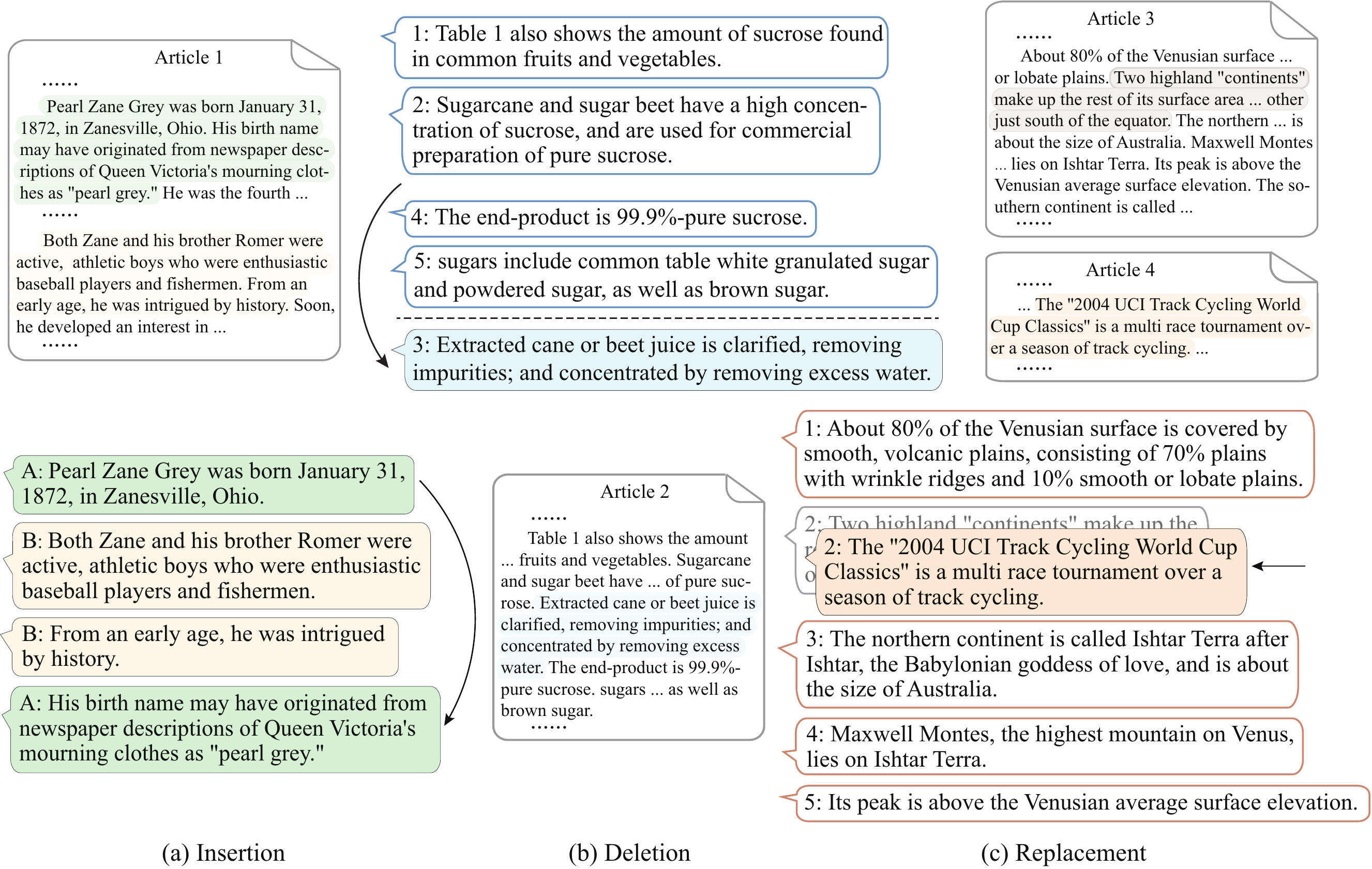}
\caption{Three dialogue-oriented pre-training strategies.}
\label{fig:Module}
\end{figure*}
For dialogue-oriented pre-training, 
we split and sample sentences as "utterances" from the general Wikipedia corpus to simulate dialogue-like features. We design three dialogue-oriented pre-training strategies (i.e., Insertion, Deletion, Replacement) to jointly learn dialogue related characteristics based on the plain PrLMs. A special token $\texttt{[SOT]}$ is added before each "utterance", which tells that it is a start of a turn and matches the  realistic scene of turn shift. The three tasks use the embedding of $\texttt{[SOT]}$ to represent each utterance and conduct targeted pre-training, which enables $\texttt{[SOT]}$ to learn dialogue related representation about speaker-awareness, continuity and consistency respectively. Figure \ref{fig:Module} shows the overview of the three strategies.
\paragraph{Insertion}
In a real scenario, the speaker role might shift or not for each turn as a conversation goes on. Two speakers may carry out a conversation in turn or one speaker may continuously shoot multiple utterances. Considering a conversation session of four sentences between speaker A and B, we consider three possible cases: AABB, ABAB ABBA. The next time A speaks will happen after 0,1, or 2 turns from the last time. To enable Dialog-PrLM aware of the speaker role information,  we should first simulate a two-party conversation on Wikipedia. We sample two continuous sentences $\{ {u_{A1}},{u_{A2}}\}$ in one paragraph of an article as the two utterances of A, and sample two continuous ones $\{ {u_{B1}},{u_{B2}}\}$ in the following paragraph of the same article as what B says. Sampling from the same article is to ensure they are talking about one topic in general, which is in line with the realistic scenario and increases the difficulty of prediction. The "continuous" sentences simulate that A continues to express his opinion after being  interrupted by B. We insert ${u_{A2}}$ into $\{ {u_{B1}},{u_{B2}}\}$, and add $\texttt{[SOT]}$ before each utterance. We will not disrupt the utterance order inside one speaker, and also keep the overall order that A first then B. In this way, we can get three cases mentioned above. One possible input case is listed here:

$X_{ins} = \texttt{[CLS]}\texttt{[SOT]}{u_{A1}}\texttt{[SOT]}{u_{B1}}\texttt{[SOT]}$ ${u_{A2}}\texttt{[SOT]}{u_{B2}}\texttt{[SEP]}$

The insertion task is to predict the next utterance of A. The whole sequence is encoded by PrLM,  and 
we calculate the cosine similarity of the $\texttt{[SOT]}$ embedding of $u_{A1}$ with the other three utterances as matching scores to predict the $u_{A2}$. These three scores are passed to a softmax layer and use cross entropy as the insertion loss $\mathcal{L}_{ins}$.
\paragraph{Deletion}
The plain pre-training tasks like MLM, NSP or SOP of PrLMs just enable the model to learn token-level or sentence-level semantic information, and they fail to catch dialogue related signals like continuity, which also helps to choose the answer that is coherent with context. We sample continuous $k$ sentences $\{ {u_1},{u_2}, \cdots ,{u_k}\}$ from one paragraph and randomly delete ${u_i}$ from the first $k-1$ (We do not choose $u_k$, as there is no $\texttt{[SOT]}$ after $u_{k-1}$). The input sequence of the PrLM is:
 
$X_{del}=\texttt{[CLS]}\texttt{[SOT]}{u_1}\cdots\texttt{[SOT]}{u_{i - 1}}$ 
$\texttt{[SOT]}{u_{i + 1}}\cdots\texttt{[SOT]}{u_k}\texttt{[SEP]}\texttt{[SOT]}{u_i}\texttt{[SEP]}$

We append $u_i$ at end and use $\texttt{[SEP]}$ for separation. Similarly, we calculate the cosine similarity of the $\texttt{[SOT]}$ embedding of $u_{i}$ with the other remaining $k-1$ utterances to predict the $\texttt{[SOT]}$ of $u_{i+1}$, where $u_{i}$ should be inserted back. These $k-1$ scores are passed to a softmax layer and use cross entropy as the deletion loss $\mathcal{L}_{del}$.
\paragraph{Replacement}
The replacement task is to make Dialog-PrLM recognize the inconsistent utterance within a dialogue session, so that it will select the proper response which is consistent with the context in both style and context. Similar to deletion, we sample continuous $k$ sentences $\{ {u_1},{u_2}, \cdots ,{u_k}\}$ from one paragraph, and then we sample one sentence $u_r$ from another article, which is used to replace a randomly chosen ${u_i}$ in $\{ {u_1},{u_2}, \cdots ,{u_k}\}$. The input sequence is:

$X_{rep}=\texttt{[CLS]}\texttt{[SOT]}{u_1}...\texttt{[SOT]}{u_{i - 1}}\texttt{[SOT]}$ ${u_r}\texttt{[SOT]}{u_{i + 1}}\cdots\texttt{[SOT]}{u_k}\texttt{[SEP]}$

Each $\texttt{[SOT]}$ is gathered after encoding, and passed to a linear layer to get a score:
\[scor{e_{{u_j}}} = {W_r}{E_j} + {b_r}\]
where $j=1...i-1, r, i+1,...k$, and $W_r, b_r$ are trainable parameters. $E_j$ is the embedding of the $j$th $\texttt{[SOT]}$. These $k$ scores are passed to a softmax layer and use cross entropy as the replacement loss $\mathcal{L}_{rep}$.

We adopt multi-task learning and define the final dialogue-oriented pre-training loss $\mathcal{L}_{gen}$ as:
\[{{\cal L}_{gen}} = {{\cal L}_{ins}} + {{\cal L}_{del}} + {{\cal L}_{rep}}\]
\section{Use of Dialogue-oriented Pre-training}
Our Dialog-PrLM may be used in terms of domain fine-funing or multi-task learning: (1) Domain fine-tuning: Our Dialog-PrLM is fine-tuned on the target response selection task. (2) Specific post-training: Our pre-training strategies are slightly adjusted and applied to specific multi-turn dialogue datasets. (3) Domain multi-task learning: the target response selection task jointly learns with the three auxiliary post-training tasks in (2) on Dialog-PrLM.
\subsection{Domain Fine-tuning}
\label{sec:domain fine-tuning}
After our dialogue-oriented pre-training, the target response selection task can be fine-tuned on our Dialog-PrLM. We denote the dataset as $D = \{ {(C,R,Y)_k}\} _{k = 1}^N$, where $C$ is dialogue context, and $R$ is the candidate response, and $Y \in \{ 0,1\}$ is the label indicating whether $R$ is a proper response for $C$. Besides, $C = \{ {U_1},{\rm{ }}...,{U_n}\}$ and ${U_i},1 \le i \le n$ is the $i$-th utterance in context $C$. We concatenate all utterances $\{ {U_{i}}\} _{i=1}^n$ as well as the response $R$ and add $\left[ \texttt{SOT} \right]$ before each to represent the following sequence: 

$X =  \left[ \texttt{CLS} \right] \left[ \texttt{SOT}\right]{U_1}\left[ \texttt{SOT} \right]{U_2}.{\rm{ }}.{\rm{ }}.\left[ \texttt{SOT} \right]{U_n}$
$\left[ \texttt{SEP} \right] \left[ \texttt{SOT} \right]R\left[ \texttt{SEP} \right]$

With the pre-training, Dialog-PrLM becomes effectively capable of representing each utterance. Therefore, rather than directly using $\left[ \texttt{CLS} \right]$, we pass all embeddings $E$ of $\left[ \texttt{SOT} \right]$s to GRU to model sequential interaction of the context and response, whose final hidden state $H$ is used for generating matching score $s$ for $C$ and $R$:
\begin{align*}
&H = {\mathop{\rm GRU}\nolimits} \left( E \right), \\
&s = {\rm{sigmoid}}(WH + b)
\end{align*}
where $E \in \mathbb{R}^{(n+1)\times{d}}, H \in \mathbb{R}^{d}$, and $W\in  \mathbb{R}^{d}, b \in \mathbb{R}$ are trainable parameters. The response selection loss ${{{\cal L}_{fine - tune}}}$ is:
\begin{align*}
&p(Y\mid C,R) = sY+(1 - s)(1 - Y)\\
&{L_{fine - tune}}{\rm{ }} =  - \frac{1}{N}\sum\limits_{(C,R,Y) \in D} {\log } (p(Y\mid C,R))
\end{align*}
\subsection{Specific Post-training}
Because the target dataset usually concentrates on a specific domain, existing works tend to introduce self-supervised post-training on the target domain in order to incorporate the in-domain knowledge. Here we can also apply the three strategies to the target multi-turn dialogue dataset as self-supervised auxiliary tasks, which jointly learn with the response selection task in Section \ref{sec:domain fine-tuning}.

To conduct the three auxiliary tasks, we should first sample from multi-turn dialogues to build post-training datasets. For the insertion task, there is a little difference from that in Wikipedia. 
We randomly choose $k$ continuous utterances $\{ {u_1},{u_2}, \cdots ,{u_k}\}$ from a dialogue, and fix $u_1$ but randomly insert $u_2$ to any interval among $\{{u_3}, \cdots ,{u_k}\}$. The input sequence is:

$X'_{ins} =  \left[ \texttt{CLS} \right] \left[ \texttt{SOT}\right]{u_1}\left[ \texttt{SOT} \right]{u_3}\cdots\left[ \texttt{SOT}\right]{u_i} $ $ \left[ \texttt{SOT}\right]{u_2}\left[ \texttt{SOT}\right]{u_{i+1}}  \cdots\left[ \texttt{SOT} \right]{u_k}\left[ \texttt{SEP} \right]$, 

We calculate the cosine similarity of the $\texttt{[SOT]}$ embedding of $u_{1}$ with the other following utterances as matching scores to predict the $u_{2}$. The loss is denoted as ${{{\cal L}'}_{ins}}$. Considering the following turn ($u_2$) tends to be more related to $u_1$ compared with the next utterance of $u_1$'s speaker (denoted as $u_t$), we do not predict $u_t$ as what we do in Wikipredia. But they are both expected to recognize the most related utterance with $u_1$, which helps select the proper response.

For the deletion and replacement tasks, we sample continuous $k$ utterances from one dialogue, and conduct deletion or replacement in the same way as Wikipedia. The post-training losses for both tasks on the target domain are denoted as ${{{\cal L}'}_{del}}, {{{\cal L}'}_{rep}}$ respectively.
%
\subsection{Domain Multi-task Learning}
\label{sec: domain multi-task learning}
We apply the multi-task learning framework on the target domain on our Dialog-PrLM. Carried with dialogue related features from the general corpus, Dialog-PrLM is expected to learn from the target domain together with the target task. We train the response selection task in the same way as \ref{sec:domain fine-tuning}, and denote the loss as ${{{\cal L}'}_{reselect}}$. The final loss is to sum up response selection loss and the three auxiliary task losses:
\[{{\cal L}_{final}} = {{{\cal L}'}_{ins}} + {{{\cal L}'}_{del}} + {{{\cal L}'}_{rep}} + {{{\cal L}'}_{reselect}}\]
\section{Implementation of ~~~~~~~~~~~~~~~~~~~~~~~~~~~~~~~~~~~~~ Dialogue-oriented Pre-training}
For dialogue-oriented pre-training, we sample train and valid datasets for the insertion, deletion and replacement tasks on both English and Chinese Wikipedia. To prevent information leakage (e.g. the model peeks at the correct utterance order from deletion samples when conducting replacement), we divide all the articles equally into three disjoint equal sets to sample from for the three tasks respectively. Data statistics are in Table \ref{tab: repretrain_statistics}.
\begin{table}[ht]
\footnotesize
\centering
\begin{tabular}{l|rr|rr}
\hline 
\multirow{2}*{\textbf{Statistics}}& \multicolumn{2}{c|}{English}&\multicolumn{2}{c}{Chinese} \\ 
~&Train&Valid&Train&Valid\\ 
\hline 
\#articles/task & 330k& 10k & 67k & 20k \\
Insertion & 1.5M & 47k & 638k & 189k\\
Deletion & 1M & 31k & 508k & 146k\\
Replacement & 1M & 32k & 542k & 160k \\
\hline
\end{tabular}
\caption{\label{tab: repretrain_statistics} Statistics for dialogue-oriented pre-training.}
\end{table}

For English Wikipedia (totally 1,060,131 articles), we sample from 330,000/10,000 articles for training/evaluation respectively for each task. To ensure data quality, we omit the "References" and "Literature" part. For insertion, we sample twice disjointly from one paragraph which has more than 4 sentences, and then sample twice from the next satisfactory one to construct two training samples, which goes on until the end of an article. For deletion and replacement, we sample $k$ continuous sentences as a training sample from each paragraph with more than $k$ sentences. We limit the maximum words of each sample to 400 to prevent overflows after tokenization.

For Chinese Wikipedia (totally 262,405 articles), we sample from 67,468/20,000 articles for training/evaluation respectively for each task. As Chinese corpus is much smaller than the English one, we sample disjoint $k$ continuous sentences as much as we can from each paragraph with more than $k$ sentences for the deletion and replacement tasks. As to insertion, we conduct sampling the same way as English. The maximum length of each sample is limited to 450.
\section{Experiments}
\subsection{Datasets}
\begin{table*}[ht]
\renewcommand\tabcolsep{0.8pt}
\footnotesize
\centering
\begin{tabular}{c|l|ccc|cccccc|ccc}
\hline 
&\multirow{2}*{\textbf{Model}}& \multicolumn{3}{c|}{E-commerce}&\multicolumn{6}{c|}{Douban}& \multicolumn{3}{c}{Ubuntu}\\ 
&~&$R_{10}@1$&$R_{10}@2$&$R_{10}@5$&MAP&MRR&P@1&$R_{10}@1$&$R_{10}@2$&$R_{10}@5$&$R_{10}@1$&$R_{10}@2$&$R_{10}@5$\\ 
\hline
&DialoGPT&-&-&-&-&-&-&-&-&-&79.0&88.5&97.1\\
&TOD-BERT&-&-&-&-&-&-&-&-&-&79.7&89.0&97.4\\
\hline
&BERT-[CLS]&62.7&82.2&96.2&58.7&62.7&45.1&27.6&45.8&82.7&81.9&90.4&97.8\\
~~$\diamondsuit$~~&BERT-[SEP]&65.1&84.8&97.4&59.5&63.9&46.0&27.7&46.9&84.3&82.1&90.5&97.8\\
~~$\diamondsuit$~~&Dialog-BERT&\textbf{66.2}&\textbf{85.5}&\textbf{97.6}&\textbf{60.0}&\textbf{64.1}&\textbf{46.9}&\textbf{28.9}&46.7&83.3&\textbf{82.3}&\textbf{90.6}&97.7\\
~~$\clubsuit$~~&BERT+multi-task&65.8&84.6&97.6&60.2&64.7&46.9&28.5&48.6&82.5&85.0&92.5&98.3\\
~~$\clubsuit$~~&Dialog-BERT+multi-task&\textbf{68.0}&\textbf{85.3}&\textbf{97.7}&\textbf{60.9}&\textbf{64.9}&\textbf{48.0}&\textbf{30.0}&47.9&\textbf{82.9}&\textbf{85.4}&\textbf{92.8}&\textbf{98.5}\\
\hline
&ELECTRA-[CLS]&58.2&79.6&96.9&59.0&63.2&44.8&27.6&47.3&82.8&82.5&90.7&97.8\\
~~$\heartsuit$~~&ELECTRA-[SEP]&60.4&80.6&96.3&58.8&62.5&44.2&26.9&46.3&84.1&82.2&90.7&97.8\\
~~$\heartsuit$~~&Dialog-ELECTRA&\textbf{61.1}&\textbf{81.4}&\textbf{96.9}&\textbf{59.8}&\textbf{64.1}&\textbf{46.5}&\textbf{28.3}&\textbf{47.7}&\textbf{84.1}&\textbf{83.5}&\textbf{91.4}&\textbf{98.0}\\
~~$\spadesuit$~~&ELECTRA+multi-task&68.1&86.8&97.9&61.4&65.3&47.5&29.6&50.6&83.8&86.6&93.4&98.5\\
~~$\spadesuit$~~&Dialog-ELECTRA+multi-task&\textbf{68.3}&86.3&\textbf{98.0}&\textbf{61.6}&\textbf{65.6}&\textbf{48.3}&\textbf{30.0}&49.8&\textbf{84.7}&\textbf{86.8}&\textbf{93.6}&\textbf{98.6}\\
\hline
\end{tabular}
\caption{\label{tab: sel_results} Response selection results on E-commerce, Douban and Ubuntu datasets. The symbols on the left indicate the corresponding comparison groups. The best results in each group are in boldface.}
\end{table*}
For the target response selection selection task, our Dialog-PrLM is fine-tuned on three widely used benchmark datasets: 
(1) \textbf{E-commerce Corpus} \citep{zhang-etal-2018-modeling}: includes conversations between customers and shopkeepers from the largest e-commerce platform Taobao 
in China. 
(2) \textbf{Douban Corpus} \citep{wu-etal-2017-sequential}: consists of multi-turn conversations from the Douban group, 
which is a popular social networking service in China. 
(3) \textbf{Ubuntu Corpus (v1.0)}\citep{lowe-etal-2015-ubuntu} consists of English multi-turn conversations about technical support collected from chat logs of the Ubuntu forum. 

As to the three auxiliary tasks for domain multi-task learning, we conduct sampling from the batch when training the response selection task. Dialogues will be neglected if they are less than 3 utterances. Different from general pre-training, we do not require every dialogue to have at least $k$ sentences for all the three tasks.

For evaluation, we use the same metric $R_n@k$ as previous works, which selects $k$ best matchable candidate responses among $n$ and calculates the recall of the true ones. 
We also use MAP (Mean Average Precision), MRR (Mean Reciprocal Rank), and Precision-at-one P@1 as previous works.
\subsection{Experimental Settings}
We work on two different PrLMs: BERT (\textit{bert-base-uncased, bert-base-chinese}) and ELECTRA (\textit{electra-base-discriminator, chinese-electra-180g-base-discriminator})\footnote{Both datasets and code are available at \url{https://github.com/xyease/Dialog-PrLM}}. 

For our dialogue-oriented pre-training on Wikipedia, the max input sequence length is set to 512 after WordPiece tokenization. We set the learning rate as 2e-5 with a warmup proportion of 10\%. The plain PrLMs are continuously pre-trained with batch size of 8 per task for BERT and 16 per task for ELECTRA.  It is trained for 1 epoch and evaluated every 10000 steps. The model with the best average accuracy of the three tasks is saved as Dialog-PrLM. The pre-training experiment needs 4 nVidia RTX 2080 GPUs.

For fine-tuning on our Dialog-PrLMs, the batch size is 32 and the max sequence length is 350. The model is trained for 5 epochs and evaluated after each epoch on the three datasets and both Dialog-PrLMs. Other settings are the same as dialogue-oriented pre-training. For domain multi-task learning on our Dialog-PrLMs, the batch size is 16, and the epoch is 3 for Douban on BERT, 4 for Ubuntu on ELECTRA and 5 for other cases. Other settings are the same as fine-tuning. The $k$ value for both pre-training and domain multi-task learning is 5. The fine-tuning/multi-task learning experiments need 1/2 nVidia RTX 2080 GPUs.
\subsection{Results}
\begin{table*}[ht]
\renewcommand\tabcolsep{3pt}
\footnotesize
\centering
\begin{tabular}{l|ccc|cccccc|ccc}
\hline 
\multirow{2}*{\textbf{Model}}& \multicolumn{3}{c|}{E-commerce}&\multicolumn{6}{c|}{Douban}& \multicolumn{3}{c}{Ubuntu}\\
~&$R_{10}@1$&$R_{10}@2$&$R_{10}@5$&MAP&MRR&P@1&$R_{10}@1$&$R_{10}@2$&$R_{10}@5$&$R_{10}@1$&$R_{10}@2$&$R_{10}@5$\\ 
\hline 
Dialog-BERT&66.2&85.5&97.6&60.0&64.1&46.9&28.9&46.7&83.3&82.3&90.6&97.7\\
\hline 
w/o Insertion&64.9&83.6&97.7&59.0&63.2&45.1&27.8&46.5&83.2&82.1&90.5&97.8\\
w/o Deletion&64.2&83.7&97.5&59.0&63.1&44.7&27.5&45.9&83.8&82.1&90.6&97.8\\
w/o
Replacement&64.4&84.7&97.6&59.8&63.6&45.4&28.4&46.8&83.6&82.1&90.4&97.8\\
\hline
\end{tabular}
\caption{\label{tab: ablation1} Ablation results for dialogue-oriented pre-training.}
\end{table*}
To verify the effectiveness of our method, we conduct extensive empirical studies on three multi-turn dialogue benchmarks. We are aware that applying complicated matching networks, speaker embeddings \citep{gu2020speaker} or other various auxiliary tasks \citep{whang2020domain,xu2021learning, whang2021ums} would achieve further improvement, but to fairly evaluate the general-purpose pre-training for dialogue tasks, we still follow the standard fine-tuning procedure on Dialog-PrLM by excluding those too advanced auxiliary techniques. 

\paragraph{BERT-[CLS]:} Each utterance $U_i$ and  the candidate response $R$ are concatenated as $\left[ \texttt{CLS} \right]{U_1}{U_2}.{\rm{ }}.{\rm{ }}.{U_n}\left[ \texttt{SEP} \right]R\left[ \texttt{SEP} \right]$ and then fed into the pla BERT model. The output embedding of $\left[ \texttt{CLS} \right]$ is used for classification.
\paragraph{BERT-[SEP]:} Rather than just use $\left[ \texttt{CLS} \right]$, we append $\left[ \texttt{SEP} \right]$ to each utterance or response to represent the previous sequence: $\left[ \texttt{CLS} \right]{U_1}$ $\left[ \texttt{SEP} \right]{U_2}.{\rm{ }}.{\rm{ }}.\left[ \texttt{SEP} \right]{U_n}\left[ \texttt{SEP} \right]R\left[ \texttt{SEP} \right]$, which is then fed into the original BERT model. The output embeddings of $\left[ \texttt{SEP} \right]$ are gathered and fed into GRU for a matching vector like Section \ref{sec:domain fine-tuning}.
\paragraph{Dialog-BERT:} This model conducts dialogue-oriented pre-training on the original BERT, we fine-tune on our Dialog-BERT through feeding $\left[ \texttt{SOT} \right]$ embeddings to GRU.
\paragraph{BERT+multi-task:} The response selection task is trained with the three auxiliary tasks on target datasets. We also add the special token $\left[ \texttt{SOT} \right]$, and the only difference from Section \ref{sec: domain multi-task learning} is that the joint learning is conducted on the original BERT.
\paragraph{Dialog-BERT+multi-task:} As described in Section \ref{sec: domain multi-task learning}, we conduct domain multi-task learning on our pre-trained Dialog-BERT.

We also conduct fine-tuning on the English Ubuntu dataset with two dialogue related models: (1) DialoGPT \citep{zhang2019dialogpt} is an extension of GPT-2 that is pre-trained on Reddit data from scratch. (2) TOD-BERT \citep{wu-etal-2020-tod} is trained on a combination of 9 task-oriented dialogue datasets over BERT and  incorporates response selection objective. Experiments on ELECTRA are conduct in the same way with BERT. The results are in Table \ref{tab: sel_results}. Below PrLM- denotes BERT or ELECTRA.

Compared to the unsatisfactory results from both DialoGPT and TOD-BERT, it demonstrates the powerfulness and universities of our proposed dialogue-oriented pre-training. Compared with PrLM-[CLS], PrLM-[SEP] performs better in general except a little decrease in Ubuntu and Douban on ELECTRA, which shows that modelling the sequential interaction of the dialogue context and response helps improve performance.  

After conducting the dialogue-oriented pre-training on Wikipredia, our Dialog-PrLM achieves further improvement on the three datasets and the two PrLMs, which shows that the three targeted training strategies enables the $\texttt{[SOT]}$ token in Dialog-PrLM to grasp dialogue related nature (e.g. speaker-awareness, continuity, consistency) at the same time, so that it is more capable of representing an utterance compared with $\texttt{[SEP]}$ in the plain PrLM (PrLM-[SEP]).

When we train the response selection task jointly with the three auxiliary tasks on target datasets, the domain multi-task learning on our Dialog-PrLM (Dialog-PrLM+multi-task) is still always performing better than on the plain PrLM (PrLM+multi-task). Having re-learned broader representation on general corpus, domain post-training further incorporates $\texttt{[SOT]}$ with dialogue related feature from in-domain multi-dialogues and thus helps choose the correct response.

Compare PrLM-[SEP] with PrLM+multi-task,  Dialog-PrLM with Dialog-PrLM+multi-task, domain multi-task learning indeed achieves improvements due to its incorporated in-domain dialogue related knowledge, which verifies the effectiveness of our proposed three strategies when applying to domain multi-turn dialogue datasets.

In conclusion, conducting dialogue related feature pre-training with our proposed three strategies on Wikipredia (Dialog-PrLM)  helps achieve improvements when fine-tuning, and it will further improve when applying these strategies to domain multi-turn dialogues (Dialog-PrLM+multi-task).
\section{Analysis}
\subsection{Ablation Study}
In order to investigate the performance of each strategy, we conduct ablation experiments for both pre-training and domain multi-task learning. Results are shown in Tables \ref{tab: ablation1} and \ref{tab: ablation2} respectively.

The results in Table \ref{tab: ablation1} indicate that the insertion, deletion and replacement tasks jointly contribute to the final increase. Influence on Ubuntu seems less than that on Douban and E-commerce, as the Ubuntu corpus contains many terminologies that do not usually appear in general corpora (e.g., apt-get, lsmod and grep) \citep{whang2020domain}, so according to BERT+multi-task in Table \ref{tab: sel_results}, conducting domain post-training is more effective. 

We also do ablation study for Douban on Dialog-BERT in Table \ref{tab: ablation2} to explore the performance of three auxiliary tasks when applying to the target multi-turn dialogue datasets. Similarly, removing any part leads to worse performance, showing the necessity of each task.
\begin{table}[ht]
\renewcommand\tabcolsep{0.8pt}
\footnotesize
\centering
\begin{tabular}{l|cccccc}
\hline 
\multirow{2}*{\textbf{Model}}&\multicolumn{5}{c}{Douban}\\
~&MAP&MRR&P@1&$R_{10}@1$&$R_{10}@2$\\
\hline 
Dialog-BERT+multi-task&60.9&64.9&48.0&30.0&47.9\\
\hline 
w/o Insertion&58.2&62.5&44.8&27.4&45.0\\
w/o Deletion&60.6&64.6&47.1&29.1&48.8\\
w/o Replacement&60.2&64.1&46.3&29.0&48.5\\
\hline
\end{tabular}
\caption{\label{tab: ablation2} Ablation for domain multi-task learning.}
\end{table}
\subsection{Utterance Representation Test}
\begin{table*}[ht]
  \footnotesize
   \renewcommand\tabcolsep{2pt}
\centering
\begin{tabular}{lccc}
\toprule
\multicolumn{1}{c}{\textbf{Dialogue Context \& Response}} & $\texttt{[CLS]}$&$\texttt{[SEP]}$&$\texttt{[SOT]}$\\ 
\midrule
\textbf{Douban} \\
1: \textit{how about going to Xitang during the Spring Festival?} & 0.8424&1 &-0.4001\\
2: \textit{{\color{SpeakerBColor}I also want to go that time}, but it will be more expensive for accommodation than weekend. }& 0.9364&1 & \textbf{0.8666}\\
3: \textit{have you contacted about where to live?} & 0.9694&1 & -0.2902\\
4: \textit{if you haven't booked accommodation, it seems very difficult to book during the Spring Festival. }& 0.9495&1 &-0.1643\\
5: \textit{will there be many people, or no shops open during the Spring Festival?} & 0.9024&1  & -0.5064\\
6: \textit{it's okay if no stores open. i enjoy quiet and just want somewhere to drink and eat.} & 0.9737&1 & -0.1644\\
7: \textit{but traffic is a hassle during the Spring Festival.} & 0.9684&1 & -0.6139\\
8: \textit{{\color{SpeakerBColor}there are several people in this group xxx who go to Xitang}, you can join them to cook together} & 0.8865&1 & \textbf{0.8840}\\
\textbf{Response:} \textit{is there anyone in Beijing, {\color{SpeakerAColor}let's go together sometime.} }& &  \\
\midrule
\textbf{Ubuntu} \\
1: \textit{hello} & 0.9279&0.9956 & -0.9791\\
2: \textit{what is the best server for update ubuntu} & 0.8027&0.9957 & -0.9653\\
3: \textit{whichever is closest to you generally they all have the same content} &0.9738&0.9949&  -0.8846\\
4: \textit{have you read the documentation for the {\color{SpeakerBColor}script}}& 0.9643&0.9966 & -0.2836\\
5: \textit{i know thats the error does the {\color{SpeakerBColor}script} s documentation indicate what modules or packages are}& \multirow{2}*{0.8176}&\multirow{2}*{0.9964} &\multirow{2}*{ \textbf{0.5233}}\\
\textit{~~~~~quried to run it} &&&\\
6:\textit{ u can tell me how to install python modules and i will install the modules in the python {\color{SpeakerBColor}script}} & 0.8940&0.9977 & \textbf{0.5287}\\
7: \textit{where did you get this {\color{SpeakerBColor}script} }& 0.9328&0.9980 & \textbf{0.5105}\\
\textbf{Response:} \textit{can you be more specific what is this {\color{SpeakerAColor}script} supposed to do import logging} & & & \\
\midrule
\textbf{E-commerce} \\
1: \textit{please help me issue an invoice with the title of xxx. }& 0.8864&1&-0.9791\\
2: \textit{okay.} & 0.8726&1&-0.8137 \\
3: \textit{ok, is it difficult to package?}& 0.9249&1&-0.6777\\
4: \textit{not at all.} & 0.7843&1&-0.9359\\
5: \textit{please send me a picture tutorial later, i will do it myself.} & 0.8648&1&-0.3668\\
6: \textit{okay, if you have any question, please feel free to consult us. }& 0.9381&1&0.3211\\
7: \textit{fine, {\color{SpeakerBColor}can you deliver it today?}}  & 0.9265&1&\textbf{0.9899}\\
8: \textit{yes, please check the {\color{SpeakerBColor}delivery} address. }& 0.9240&1&\textbf{0.7117}\\
9: \textit{correct. }& 0.6862&1&\textbf{0.5704} \\
10: \textit{ok, if pay before 16:00 today, {\color{SpeakerBColor}it will be delivered today, otherwise it will be delivered next day.}} &0.9550&1&\textbf{0.5866}\\
11: \textit{i've already {\color{SpeakerBColor}paid}.} & 0.8663&1&\textbf{0.0622}\\
\textbf{Response:} \textit{ok, {\color{SpeakerAColor}we will deliver the goods as soon as possible today}, please wait at patience. }&  \\
\arrayrulecolor{black} \hline
\end{tabular}
\caption{\label{tab:utt repre test} Examples from test sets of Douban, Ubuntu and E-commerce respectively. Similarity scores $\ge $ 0.5 of our Dialog-BERT($\texttt{[SOT]}$) are bold, indicating the corresponding utterances are considered relevant to the response.}
   \end{table*}
We have added a special token \texttt{[SOT]} before each utterance or response to represent the following sequence. After pre-training on Wikipedia on PrLMs, \texttt{[SOT]} of our Dialog-PrLM is expected to obtain the respective utterance representation through the three dialogue-oriented strategies. 

To explore the semantic information of \texttt{[SOT]} of our Dialog-BERT, we calculate the cosine similarity of the correct response to each utterance in the dialogue context ($\texttt{[SOT]}$). Table \ref{tab:utt repre test} lists examples from the three target datasets respectively. For comparison, we use BERT to encode each utterance or response and use $\texttt{[CLS]}$ for calculation ($\texttt{[CLS]}$). We also concatenate utterances and response with separation of $\texttt{[SEP]}$ on BERT and then split to calculate similarity ($\texttt{[SEP]}$).

We observe that for all examples,  both BERT($\texttt{[CLS]}$) and BERT($\texttt{[SEP]}$) can not discriminate which utterance is related with the correct answer. All the utterances are treated the same way including the irrelevant ones, which leads to much noise for response selection. 

After conducting the dialogue-oriented pre-training on Wikipedia, Dialog-BERT learns a stark sense of "irrelevant" and "relevant". It is able to concentrate on the most critical utterances and distinguish from the irrelevant ones by a large margin. For the example of Douban, Dialog-BERT realizes that the second and last utterance are most relevant. The response asks for someone to travel together, and is related to the second utterance which expresses a wish to go and the last one which gives a group to travel together. Dialog-BERT ignores the noise about accommodation and transportation, and is able to select related utterances among noise rather than just use the last one. For Ubuntu,  Dialog-BERT concentrates on utterances about script and ignores the previous background information. For E-commerce, it recognizes the related last few utterances about delivery, and ignores the packaging information before. From examples from target datasets, our Dialog-BERT has absorbed related knowledge from our proposed dialog-oriented pre-training. The $\texttt{[SOT]}$ could better represent each utterance, which can be utilized in tasks about representation like multi-party dialogue disentanglement.  
\section{Conclusion}
This paper presents a novel general-purpose solution for dialogue tasks with pre-trained language models. To fill the gap between a detailed task and the LM task of PrLM, we propose dialogue-oriented pre-training on large scale of artificially built dialogue data which lets the resulted Dialog-PrLM enjoy both merits of general-purpose and capturing key dialogue related features including speak awareness, continuity and consistence. Our models are evaluated on three benchmark response selection datasets and achieve consistent performance improvement over the plain PrLMs.
\bibliographystyle{acl_natbib}
\bibliography{acl2021}


\end{document}